\documentclass[10pt,twocolumn,letterpaper,table,xcdraw]{article}
\pdfoutput=1
\usepackage{iccv}
\usepackage{times}
\usepackage{epsfig}
\usepackage{graphicx}
\usepackage{amsmath}
\usepackage{amssymb}
\usepackage{booktabs}
\usepackage{multirow}
\usepackage{enumerate}
\usepackage{xcolor}
\usepackage{diagbox}
\usepackage[sort]{natbib} 
\setcitestyle{numbers,square}
\usepackage[ruled,linesnumbered]{algorithm2e}
\usepackage{authblk}

\usepackage[breaklinks=true,bookmarks=false]{hyperref}

\iccvfinalcopy % *** Uncomment this line for the final submission

\def\iccvPaperID{5396} % *** Enter the ICCV Paper ID here

\ificcvfinal\fi

\begin{document}

%%%%%%%%% TITLE
\title{Unsupervised Self-Driving Attention Prediction\\ via Uncertainty Mining and Knowledge Embedding}

\author[1,2]{Pengfei Zhu}
\author[1,2]{Mengshi Qi \thanks{Corresponding author.}}
\author[2]{Xia Li}
\author[3]{Weijian Li}
\author[1,2]{Huadong Ma}
\affil[1]{Beijing Key Laboratory of Intelligent Telecommunications Software and Multimedia}
\affil[2]{Beijing University of Posts and Telecommunications}
\affil[3]{Department of Computer Science, University of Rochester \authorcr \tt \small \{zhupengfei2000,qms,mhd\}@bupt.edu.cn; lixia@bupt.cn; wli69@cs.rochester.edu}

\maketitle
% Remove page # from the first page of camera-ready.
\ificcvfinal\thispagestyle{empty}\fi

%%%%%%%%% ABSTRACT
\begin{abstract}
   Predicting attention regions of interest is an important yet challenging task for self-driving systems. Existing methodologies rely on large-scale labeled traffic datasets that are labor-intensive to obtain. Besides, the huge domain gap between natural scenes and traffic scenes in current datasets also limits the potential for model training. To address these challenges, we are the first to introduce an unsupervised way to predict self-driving attention by uncertainty modeling and driving knowledge integration. Our approach's Uncertainty Mining Branch (UMB) discovers commonalities and differences from multiple generated pseudo-labels achieved from models pre-trained on natural scenes by actively measuring the uncertainty. Meanwhile, our Knowledge Embedding Block (KEB) bridges the domain gap by incorporating driving knowledge to adaptively refine the generated pseudo-labels. Quantitative and qualitative results with equivalent or even more impressive performance compared to fully-supervised state-of-the-art approaches across all three public datasets demonstrate the effectiveness of the proposed method and the potential of this direction. The code will be made publicly available.
\end{abstract}

%%%%%%%%% BODY TEXT
\section{Introduction}

With the huge development of autonomous driving, predicting attention regions for self-driving systems~\cite{baee2021medirl,pal2020looking} has drawn rapid interest in the community. The predicted attention region provides rich contextual information to assist autonomous driving systems by locating salient areas in the traffic scene~\cite{xu2020explainable,qi2020stc,qi2019ke}. Most importantly, these salient areas are always the riskiest areas, where small perception errors can cause great harm to drive safety~\cite{kendall2017uncertainties}. Therefore, with a successful attention area prediction, computation resources can be reallocated to enhance the perception accuracy in these fatal areas to reduce driving risks, as well as increase the explainability and improve the reliability of autonomous driving systems~\cite{kim2018textual}.

\begin{figure}
    \centering
    \includegraphics[width=1\linewidth]{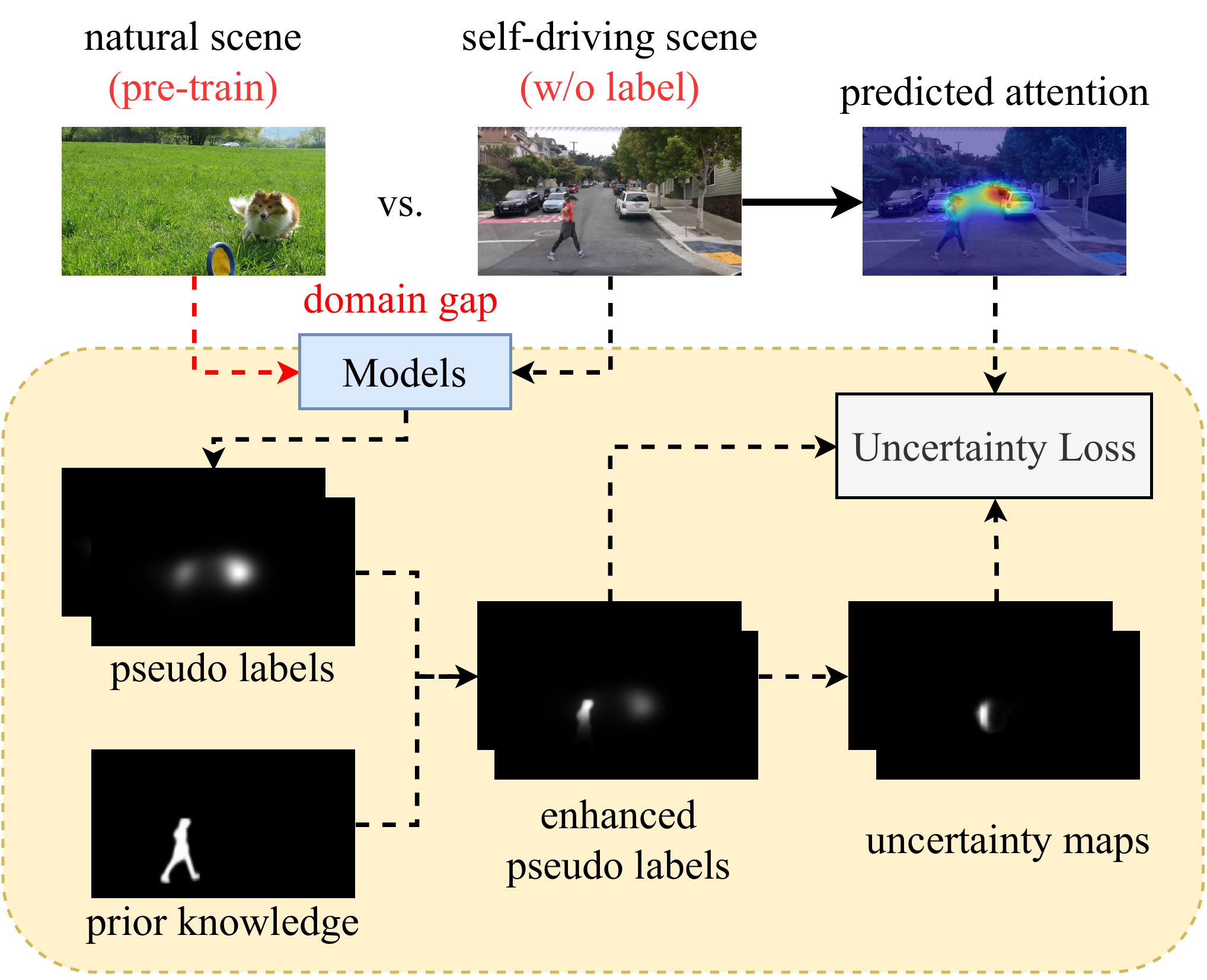}
    \caption{Illustration of the proposed unsupervised self-driving attention prediction model. Instead of relying on the ground truth labels provided by traffic datasets, our method only uses pseudo-labels generated from models pre-trained on natural scenes, and then refined the results by uncertainty mining and knowledge embedding. The red dashed line corresponds to the pre-training stage, the black dashed line refers to the training process, and the black solid line means the testing process.}
    \label{fig:motivation}
    \vspace{-3mm}
\end{figure}

Numerous datasets~\cite{xia2018predicting,alletto2016dr,fang2021dada} and methods~\cite{palazzi2018predicting,xia2018predicting,baee2021medirl,lv2020improving,kim2020advisable,lateef2021saliency} have been proposed to address self-driving attention prediction task. Though achieving encouraging performance, these methods are trained in fully-supervised ways on large-scale labeled datasets which are hard to build and unreliable. For example, one of the widely-used datasets in self-driving named DR(eye)VE~\cite{alletto2016dr} was collected in two months, by recording eight drivers taking turns driving on the same route to obtain fixation data. However, simply averaging the attention of eight drivers into one driving video will lead to the wrong attention target. Another common difficulty is the huge mismatch between the collected data and real-world environments. Another self-driving dataset BDD-A~\cite{xia2018predicting} was constructed by asking 45 participants to watch the same recorded video and imagine themselves as the drivers. But, these simulated virtual environments inevitably brought inconsistencies to real-world conditions for human labeling. Therefore, current fully-supervised methods suffer from potential biases in public datasets and then are too hard to extend to new environments. Furthermore, large-scale pre-trained models~\cite{bommasani2021opportunities} have already demonstrated strong capability in representation learning, which can beneficial to lots of downstream tasks. But how to bridge the domain gap between the specific situation (\eg self-driving scenes) and the common data pre-trained model used (\eg natural scenes) is still a challenge.

To address the above-mentioned issues, we propose a novel unsupervised framework to self-driving attention prediction, which means \textbf{1)} we do not use any ground-truth labels given by self-driving datasets, \textbf{2)} we only use pseudo-labels generated from models pre-trained on natural scene datasets. Specifically, our proposed model is achieved with two newly-designed parts: 
an uncertainty mining branch is proposed to exploit pseudo-labels' uncertainties by aligning the various distributions and thus make the result reliable; another is a knowledge embedding block which is introduced to transfer the traffic knowledge into the natural domain by segmenting the focal traffic objects with Mask-RCNN~\cite{he2017mask} pre-trained on MS-COCO~\cite{lin2014microsoft} and then enhance each pseudo-label's attention region. 

In summary, our contributions can be listed as follows:
\par\textbf{(1)} We propose a novel unsupervised framework to predict self-driving attention regions, which is not relying on any labels on traffic datasets. To the best of our knowledge, this is the first work to introduce such an unsupervised method to this specific task.
\par\textbf{(2)} We introduce an uncertainty mining branch to produce highly plausible attention maps by estimating the commonality and distinction between multiple easily obtained pseudo-labels from models pre-trained on natural scenes.
\par\textbf{(3)} We design a knowledge embedding block by incorporating rich driving knowledge to refine the produced pseudo-labels, which bridges the domain gap between autonomous driving and common domains (\eg natural scene, daily life, and sports scene).
\par\textbf{(4)} Extensive experiments on three public benchmarks with comparable or even better results compared with fully-supervised state-of-the-art approaches demonstrate the effectiveness and superiority of the proposed method.

\section{Related Work}

\begin{figure*}
    \centering
    \includegraphics[width=0.9\linewidth]{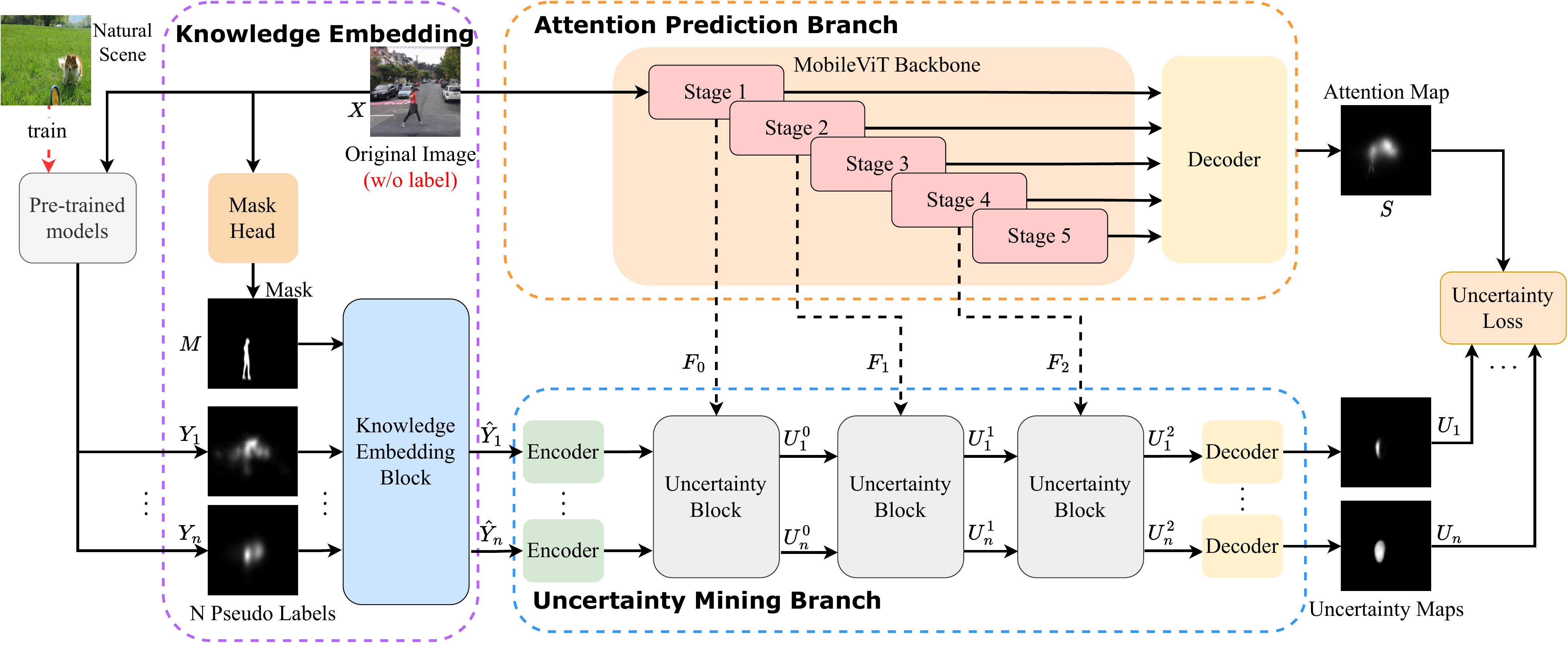}
    \caption{An overview of our proposed unsupervised self-driving attention prediction model. Our approach leverages pseudo-labels generated from models pre-trained on natural scene datasets for unsupervised training. To introduce additional semantic information for the self-driving scenario, we propose a Knowledge Embedding Block (KEB). Meanwhile, the Attention Prediction Block (APB) comprises five stages for image feature extraction, with each stage producing features subsequently fed to the decoder. Note that features extracted in stages 1, 2, and 4 are sent to three Uncertainty blocks for multi-scale feature fusion. Our Uncertainty Mining Block (UMB) employs multiple pseudo-labels with multi-scale features for fusion and mining to generate an uncertainty map for each pseudo-label. Finally, we optimize the network structure using uncertainty loss. }
    \label{fig:overview}
    \vspace{-3mm}
\end{figure*} 

\noindent{\bf Self-Driving Attention Prediction.}~With the rise of deep learning, several attempts~\cite{xia2018predicting,palazzi2018predicting,lv2020improving,fang2021dada} have been made to introduce various deep learning methods into the field of self-driving attention prediction. Palazzi \etal~\cite{palazzi2018predicting} employed a multi-branch video understanding method to predict the driver's attention in a hierarchical manner from coarse to fine. Xia \etal~\cite{xia2018predicting} addressed the center bias problem in attention prediction by assigning varying weights to each training sample based on the KL divergence between the attention map and the average attention map. Meanwhile, Baee \etal~\cite{baee2021medirl} leveraged an inverse reinforcement learning (IRL) approach to improve the accuracy of attention prediction by incorporating task-specific information. All previous studies relied on large-scale in-lab or in-car annotated datasets~\cite{fang2021dada,xia2018predicting,alletto2016dr}. DR(eye)VE~\cite{alletto2016dr} presented an in-car dataset that includes dozens of segments, which record driver's attention changes during prolonged driving in the car. BDD-A~\cite{xia2018predicting} and DADA-2000~\cite{fang2021dada} are presented as in-lab datasets that synthesize attention changes of several volunteers, providing more than 1000 clips, containing both normal and multiple emergent driving situations. To overcome the unreliable dependency of self-driving datasets, our model is the first to address self-driving attention prediction in an unsupervised manner by leveraging pseudo-labels generated by models pre-trained on natural scenes. 

%%We leverage several pseudo-labels generated by models trained on natural scene attention prediction datasets to train our attention prediction network, thereby reducing the impact of the domain gap.

%%Besides, there have been several efforts~\cite{deng2019drivers,lateef2021saliency,pal2020looking,rasouli2017ICCVW} to collect and open source large-scale self-driving attention datasets.

\noindent{\bf Saliency Detection.}~Predicted saliency regions in images or videos~\cite{jiang2018deepvs,min2019tased,droste2020unified} can approximate human’s visual attention. It has been used to evaluate the explainability of deep models~\cite{kim2018textual,xu2020explainable} and to assist other tasks, i.e., photo cropping~\cite{wang2017deep}, scene understanding~\cite{qi2019attentive,qi2020few,qi2021semantics,qi2018stagnet} and object segmentation~\cite{xu2020explainable}. However, most existing datasets~\cite{jiang2015salicon,soomro2014action,wang2018revisiting,borji2015cat2000} and methods~\cite{min2019tased,cornia2016deep,wang2021semantic,cornia2018predicting,droste2020unified,zhang2018saliency,wang2021semantic,jiang2018deepvs} are mainly focusing on natural scenes or common objects, not specially tailored into self-driving scenarios. In this work, we propose an uncertainty mining branch and a knowledge embedding strategy to bridge the domain gap between natural scenes and self-driving situations.

\noindent{\bf Uncertainty Estimation.}~Early uncertainty estimation works in deep learning mainly focus on model uncertainty, which is crucial for evaluating the accuracy and robustness of the model. A pioneer work is that Gal and Ghahramani~\cite{gal2016dropout,kendall2015bayesian} use dropout to represent model uncertainty. Lately, Kendall \etal~\cite{kendall2017uncertainties} constructs a new loss that combines data uncertainty and model uncertainty for multi-task learning~\cite{kendall2018multi}. Nowadays, uncertainty methods have been widely used in various autonomous driving tasks such as target detection~\cite{choi2019gaussian,loquercio2020general}, motion prediction~\cite{djuric2020uncertainty,feng2018towards}, semantic segmentation~\cite{blum2019fishyscapes,wang2022multi}, and etc. In the field of self-driving attention prediction, there has been no prior work that incorporates uncertainty estimation. We are the first to introduce an uncertainty mining branch to estimate the commonality and distinction between multiple pseudo-labels, and then produce plausible attention maps.
%%Kendall \etal~\cite{kendall2015bayesian} leverage stochastic dropout to model uncertainty in deep convolutional encoder-decoder architectures for scene understanding. Loquercio \etal~\cite{loquercio2020general} proposed a framework for uncertainty estimation in deep learning for autonomous driving.

\section{Method}

\subsection{Overview}

 Figure~\ref{fig:overview} shows an overview of our proposed unsupervised driving attention prediction network. Our network consists of an Attention Prediction Branch (APB), an Uncertainty Mining Branch (UMB) as well as a Knowledge Embedding Block (KEB). 
 
 Our method learns to predict self-driving attention in an unsupervised way. To achieve unsupervised learning, a naive way is to train the model with the generated pseudo-labels from a single source  model pre-trained on natural scenes. However, the large domain gap between natural environments and self-driving scenes brings strong uncertainty. Meanwhile, each single source label from a specific domain shall correspond to a different distribution, in which some particular areas may lead to strong uncertainty. Encouraged by the recent development of uncertainty estimation, we propose to improve the accuracy and robustness of our prediction by modeling uncertainty from multi-source pseudo-labels. Through the evaluation of uncertainties across various distributions, we can effectively alleviate potential discrepancies and inconsistencies. Moreover, since the generated pseudo-labels we used are directly transferred from the natural domain, they lack relevant knowledge of autonomous driving scenarios. Thus, we perform a knowledge enhancement pre-processing operation in KEB on each input pseudo-label to improve prediction results.

\noindent\textbf{Problem Formulation.}~Given an RGB input frame $X\in \mathbb{R}^{H\times W\times 3}$, APB extracts pyramid features in five levels and passes the features $F$ from the 1st, 2nd, and 4th stages as $\{F^0, F^1, F^2\}$ to explore pseudo-labels' uncertainty in UMB. APB follows the structure of U-Net~\cite{ronneberger2015u}, feeds the extracted features from the last layer into the decoder and concatenates them with the features at corresponding granularity, and outputs the final attention prediction result as $S\in {{\mathbb{R}}^{H\times W\times 1}}$ through a Readout module. In addition, before feeding pseudo-labels into UMB, we perform a knowledge enhancement process to get pseudo-labels adapted to autonomous driving scenarios with an off-the-shelf Mask Head. Then, UMB takes $N$ knowledge-embedded pseudo-labels $\hat{Y}=\left\{ {\hat{Y}_{1}},\cdots, {\hat{Y}_{n}} \right\}$ as input and estimate the uncertainty maps correspondingly, which have the same size with the final output attention map $S$. These pseudo-labels are fused with three different levels of features from APB to output the uncertainty maps $U=\left\{ {{U}_{1}},\cdots,{{U}_{N}} \right\}$. Finally, the model is trained by optimizing the uncertainty loss between the attention map and the uncertainty map.

\subsection{Uncertainty Mining Branch (UMB)}

In our work, UMB is introduced to mine the uncertainty from multi-source pseudo-labels that are generated from multiple pre-trained models. Notice that these models are pre-trained on natural scenes, not self-driving, i.e. ML-Net~\cite{cornia2016deep}, SAM~\cite{cornia2018predicting}, and UNISAL~\cite{droste2020unified} are pre-trained on SALICON~\cite{jiang2015salicon}, while TASED-Net~\cite{min2019tased} is pre-trained on DHF-1K~\cite{wang2018revisiting}. As is shown in Figure~\ref{fig:UE}, the Uncertainty Block is proposed to exchange information between pseudo-labels and multi-scale features extracted by APB, which consists of the non-local self-attention operations and merge/split mechanism~\cite{wang2018non,wang2022multi}. In our UMB, we adopt three such blocks to gather information from both pseudo-labels and multi-scale image features and enable long-range interactions among pixels. For more details please see our supplementary materials. 

Specifically, in the uncertainty block, for the $n$-th knowledge-embedded pseudo-label $\hat{Y_n}\in {{\mathbb{R}}^{H\times W\times 1}}$, we first pass it through a convolutional layer and a downsampling layer, resulting in $\frac{1}{4}$ of the original size. Then we feed it into a residual block~\cite{he2016deep} to exchange information with pseudo-labels and features maps from other sources at the same stage. The obtained results are concatenated with the input multi-source pseudo-labels and then are passed through the non-local self-attention to obtain a coarse uncertainty map $U_n$ corresponding to the $n$-{th} pseudo label, formulated as:

\begin{figure}
    \centering
    \includegraphics[width=1\linewidth]{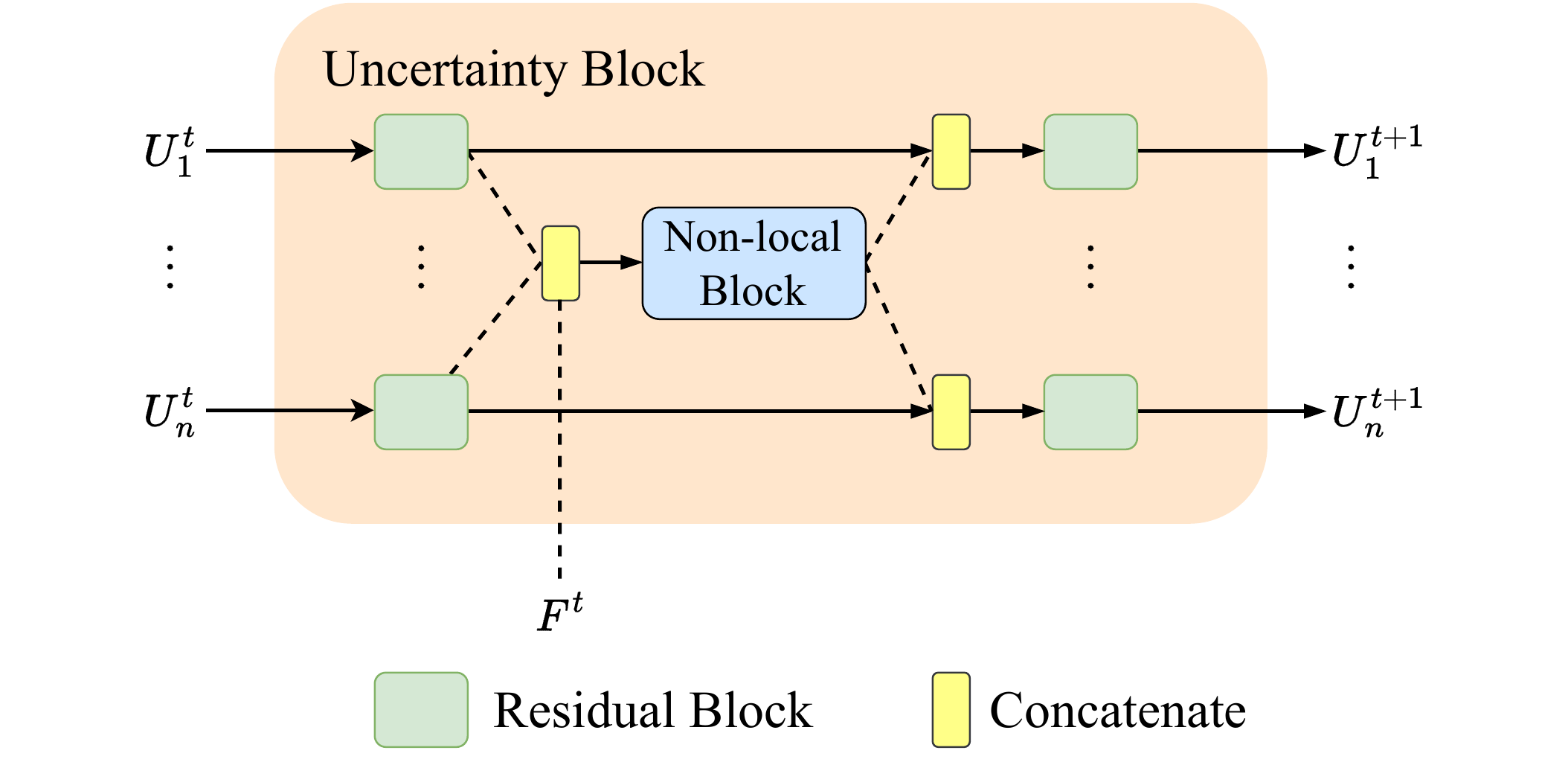}
    \caption{Illustration of the  proposed Uncertainty Block. In each stage, the input uncertainty maps $U$s from the previous stage pass through a residual block, then are concatenated with another uncertainty map and are fed into the Non-local Block. The results are concatenated with the original uncertainty map and are passed through a residual block as the input of the next stage.}
    \label{fig:UE}
    \vspace{-3mm}
\end{figure}

\begin{equation}
    {{U}_{n}^{0}}={{f}_{attn}^0}\left( \mathrm{Concat}\left( \hat{Y_{1}},\cdots ,\hat{Y_{n}},{{F}^{0}} \right) \right)+\hat{Y_{n}},
\end{equation}

\noindent where the superscripts denote the stage index, and $f_{attn}^{t}(\cdot)$ refers to non-local self-attention. Then we gradually refine ${U}_{n}^{0}$ to ${U}_{n}^{t+1}$ as follows:

\begin{equation}
    {{U}_{n}^{t+1}}={{f}_{attn}^t}\left( \mathrm{Concat}\left( U_{1}^{t},\cdots ,U_{N}^{t},{{F}^{t}} \right) \right)+U_{n}^{t}.
\end{equation}

Finally, through three uncertainty blocks, the fine-grained uncertainty map ${{U}_{n}^{2}}\in {{\mathbb{R}}^{\frac{H}{4}\times \frac{W}{4}\times 1}}$ can be obtained and then be upsampled to $U_{n}\in {{\mathbb{R}}^{H\times W\times 1}}$ in the decoder as the same size as the original input. 

%% Based on the results of the ablation experiments conducted in Section 4.4, we ultimately selected the pseudo-labels generated by ML-Net and UNISAL for use in our final training procedure.

\subsection{Knowledge Embedding Block (KEB)}
\begin{figure}
    \centering
    \includegraphics[width=1\linewidth]{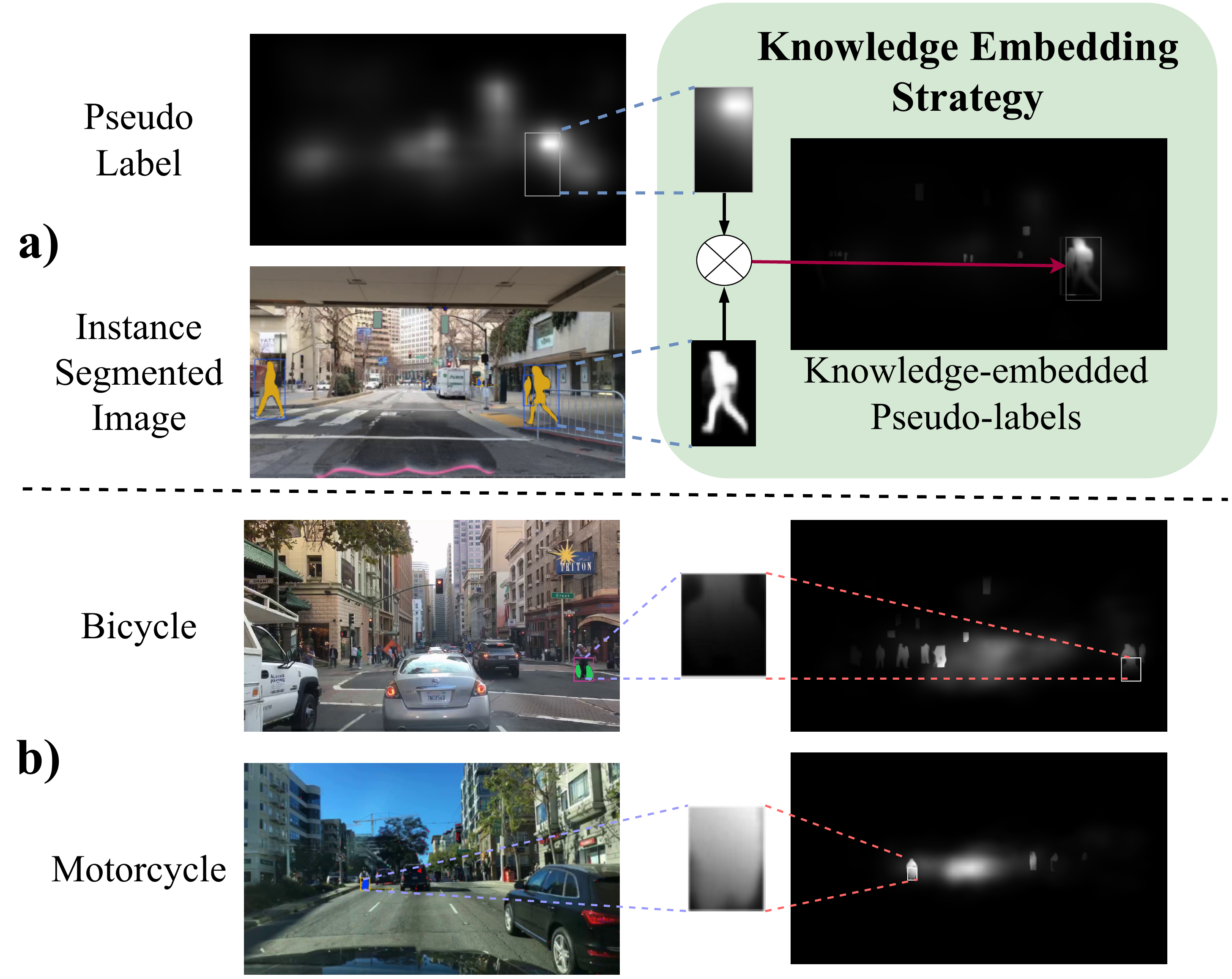}
    \caption{Illustration of the knowledge embedding strategy: a) the process of knowledge embedding for a single pseudo-label, where the salient region can be enhanced by adding the self-driving-related instance (\eg pedestrian) where the operator $\otimes$ means the operation in Eq.~\ref{eqo:prior}; b) two other examples of knowledge embedding for bicycles and motorcycles.}
    \label{fig:kes}
    \vspace{-3mm}
\end{figure}

With prior knowledge, human are able to disambiguate and discover relevant objects centered at the visual clutter~\cite{katsuki2014bottom} in visually complex scenes. Inspired by these findings, we design KEB to enhance prior driving knowledge and bridge the domain gap between natural scenes and self-driving environments. To be specific, we use the off-the-shelf Mask R-CNN pre-trained on the MS-COCO dataset~\cite{lin2014microsoft} to segment the most representative traffic objects as prior knowledge, \emph{i.e.,} pedestrians, signals, bicycles, motorcycles, and traffic signs (\eg., stop signs, road signs, etc.). During the knowledge embedding, we freeze the parameters of Mask R-CNN with the open-source checkpoints to make the knowledge embedding process practically unsupervised. Through the segmenting of the input frame with Mask-RCNN, we merge the obtained masks of different categories into a single binary mask map. Note that we explore two strategies to embed prior knowledge into different pseudo-labels: 1) concatenating them at the channel dimension and 2) fusing them to a one-channel segmentation map. For the first strategy, each pseudo-label are concatenated with the binary mask and then fed into UMB, allowing the model to learn the relationship adaptively. For the second strategy, we compose each pseudo-label with the binary mask using the following formulation:
\begin{equation}
    {{\hat{Y}}_{n}} = {{Y}_{n}}\cdot \left( {M}+\alpha  \right),
\label{eqo:prior}
\end{equation}
where $\alpha$ is a hyper-parameter that is empirically set to 0.3, $Y_{n}$ denotes the $n$-th pseudo-label, and $M$ denotes the segmented map of the corresponding image. We adopt the second strategy in our approach for better performance (for more experimental results please refer to Sec~\ref{subsec:abl}).

%%Inspired by this, we use most representative objects as additional knowledge, \emph{i.e.,} pedestrians, signals, bicycles, motorcycles, and traffic signs (\eg., stop signs, road signs, etc.). The classical instance segmentation method Mask-RCNN~\cite{he2017mask} is used to extract masks for the above-mentioned important instances. We use Mask R-CNN pre-trained on the MS-COCO dataset~\cite{lin2014microsoft}, which does not contain any pre-defined semantic information about the self-driving scenario. 

\subsection{Loss Function}

We treat the predicted attention map $S$ as a distribution over the spatial dimension and we need to normalize the generated pseudo-labels accordingly. 
% Besides, outputs from APB will go through a softmax layer to satisfy the above requirements. 
To satisfy this requirement, we apply a spatial softmax layer after APB.
Inspired by the uncertainty loss in ~\cite{kendall2017uncertainties}, we assume a Boltzmann distribution under the Bayesian theory for each pseudo-label map $\hat{Y}_n \in \mathbb{R}^{H \times W \times 1}$. Therefore, the probability of the final prediction $S$ with respect to the label $\hat{Y}_{n}$ can be calculated as follows:
\begin{equation}
    p(\hat{Y}_n | S, u_n) = \prod_i \mathrm{Softmax} (\frac{S_i}{u_n^2}),
    % p\left( y~\text{ }\!\!|\!\!\text{ }~{{f}^{w}}\left( x \right),\sigma  \right)=\mathrm{Softmax}\left( \frac{1}{{{\sigma }^{2}}}{{f}^{w}}\left( x \right) \right),
\label{eqo:base}
\end{equation}
where $u_n = 1/(H\times W)\sum_{i}^{H\times W} U_n^i$ is the final uncertainty estimation for the $n$-th pseudo-label, $i$ denotes the pixel index of $S$. Also, $u_n$ can be regarded as the temperature parameter whose magnitude determines how ‘uniform’ (flat) the distribution is. The negative log-likelihood of the whole pseudo-label map is calculated as:
\begin{equation}
    \begin{aligned}
        & -\log p(\hat{Y}_n | S, u_n) \\
        =& -\sum_i\frac{S_i}{u_n^2} + \log \sum_i \exp (\frac{S_i}{u_n^2}) \\
        \approx& ~\frac{L_{\mathrm{CE}}(S, \hat{Y}_n)}{u_n^2} + \log(u_n),
    \end{aligned}
    \label{eqo:origin}
\end{equation}
where $L_{\mathrm{CE}}(S, \hat{Y}_n)$ denotes the spatial cross entropy loss. In practice, we can instead predict the log variance $e_n = \log(u_n)^2$ to increase the numerical stability~\cite{kendall2018multi} during the training process. Now, the loss can be re-formulated as follows:
\begin{equation}
    L(S, u_n, \hat{Y}_n) = L_{\mathrm{CE}}(S, \hat{Y}_n) \cdot \exp(-e_n) + \frac{1}{2}e_n.
    \label{eq:uncertainty}
\end{equation}

Besides, we can reformulate the cross-entropy loss $L_{\mathrm{CE}}(S, Y_n)$ as follows:
\begin{equation}
    \begin{aligned}
        L_{\mathrm{CE}}(S, \hat{Y}_n) &= -\sum_i \hat{Y}_{n,i} \log(S_i) \\
        &= -\sum_i \hat{Y}_{n,i} \log(S_i) + H(\hat{Y}_n) - H(\hat{Y}_n) \\
        &= \sum_i \hat{Y}_{n,i} (\log(\hat{Y}_{n,i}) - \log(S_i)) - H(\hat{Y}_n) \\
        &= L_{\mathrm{KLD}}(\hat{Y}_n, S) - H(\hat{Y}_n),
    \end{aligned}
    \label{eq:kld}
\end{equation}
where $L_{\mathrm{KLD}}(\hat{Y}_n, S) = \sum_i \hat{Y}_{n, i} (\log(\hat{Y}_{n, i}) - \log(S_i))$ is the KL-divergence between the pseudo-label distribution and the predicted attention map distribution.
$H(\hat{Y}_n)$ is the information entropy of the distribution $\hat{Y}_n$, which is non-related to the optimization and thus can be regarded as a constant. Therefore, according to Eq.~\ref{eq:uncertainty} and Eq.~\ref{eq:kld}, and extending the calculation to all N-source pseudo-labels, we obtain the final loss as:
\begin{equation}
    L = \sum^N_{n=1} \{ L_{\mathrm{KLD}}(\hat{Y}_n, S) \cdot \exp(-e_n) + \frac{1}{2}e_n \}.
\end{equation}

Notice that our KLD uncertainty loss differs from formulas of~\cite{kullback1968information} that we assume a spatial distribution instead of a single per-channel counterpart. This assumption is crucial for derivation of Eq.~\ref{eq:kld}. For more details about the whole algorithm please refer to our supplementary material.

\section{Experimental Results}

\begin{table*}
\centering
\setlength\tabcolsep{2.5mm}
\begin{tabular}{lcccccc}
\hline
\rowcolor{gray!20}
\multicolumn{1}{c|}{} &
  \multicolumn{2}{c|}{\textbf{BDD-A}~\cite{xia2018predicting}} &
  \multicolumn{2}{c|}{\textbf{DR(eye)VE}~\cite{alletto2016dr}} &
  \multicolumn{2}{c}{\textbf{DADA-2000~\cite{fang2021dada}}} \\ \cline{2-7}
\rowcolor{gray!20}
\multicolumn{1}{c|}{\multirow{-2}{*}{\textbf{Methods}}} &
  KLD↓ &
  \multicolumn{1}{c|}{CC↑} &
  KLD↓ &
  \multicolumn{1}{c|}{CC↑} &
  KLD↓ &
  CC↑ \\ \hline
% \multicolumn{7}{c}{Supervised} \\ \hline
\multicolumn{1}{l|}{Multi-Branch~\cite{palazzi2018predicting}} &
  1.28 &
  \multicolumn{1}{c|}{0.58} &
  {\textbf{1.40}} &
  \multicolumn{1}{c|}{{\textbf{0.56}}} &
  2.27 &
  0.45 \\
\multicolumn{1}{l|}{HWS~\cite{xia2018predicting}} &
  1.34 &
  \multicolumn{1}{c|}{0.54} &
  2.12 &
  \multicolumn{1}{c|}{{0.51}} &
  2.50 &
  0.40 \\
\multicolumn{1}{l|}{SAM~\cite{cornia2018predicting}} &
  2.46 &
  \multicolumn{1}{c|}{0.25} &
  2.56 &
  \multicolumn{1}{c|}{0.38} &
  2.85 &
  0.27 \\
\multicolumn{1}{l|}{Tased-Net~\cite{min2019tased}} &
  1.79 &
  \multicolumn{1}{c|}{0.52} &
  {\underline{1.88}} &
  \multicolumn{1}{c|}{0.47} &
  {\underline{1.88}} &
  {\underline{0.53}} \\
\multicolumn{1}{l|}{MEDIRL~\cite{baee2021medirl}} &
  2.51 &
  \multicolumn{1}{c|}{{\textbf{0.74}}} &
  - &
  \multicolumn{1}{c|}{-} &
  2.93 &
  {\textbf{0.63}} \\
\multicolumn{1}{l|}{ML-Net~\cite{cornia2016deep}} &
  1.20 &
%   \multicolumn{1}{c|}{{\color[HTML]{5B9BD5} \textbf{0.64}}} &
    \multicolumn{1}{c|}{{0.64}} &
  2.00 &
  \multicolumn{1}{c|}{0.44} &
  - &
  - \\
\multicolumn{1}{l|}{UNISAL~\cite{droste2020unified}} &
  1.49 &
  \multicolumn{1}{c|}{0.58} &
  - &
  \multicolumn{1}{c|}{-} &
  - &
  - \\
\multicolumn{1}{l|}{PiCANet~\cite{liu2018picanet}} &
  {\underline{1.11}} &
  \multicolumn{1}{c|}{{0.64}} &
  - &
  \multicolumn{1}{c|}{-} &
  - &
  - \\
\multicolumn{1}{l|}{DADA~\cite{fang2021dada}} &
  - &
  \multicolumn{1}{c|}{-} &
  - &
  \multicolumn{1}{c|}{-} &
  2.19 &
  0.50 \\ 
%   \hline
% \multicolumn{7}{c}{Unsupervised} \\ \hline
\multicolumn{1}{l|}{\textbf{Ours (unsupervised)}} &
  {\textbf{1.099±0.016}} &
  \multicolumn{1}{c|}{\underline{0.640±0.007}} &
  1.901±0.004 &
  \multicolumn{1}{c|}{{\underline{0.510±0.005}}} &
  {\textbf{1.677±0.007}} &
  0.488±0.002 \\ 
  \hline
\end{tabular}
\vspace{2mm}
\caption{Performance comparison between our proposed unsupervised method and state-of-the-art fully-supervised methods. It is worth noting that our unsupervised method achieves comparable or even better performance compared with the fully-supervised methods. The numbers in bold denote the best results, and those marked with underlines denote the second best.}
\label{table:overview}
\end{table*}

\begin{table*}
\centering
\setlength\tabcolsep{1.7mm}
\begin{tabular}{c|cc|cc|cc}
\hline
\rowcolor{gray!20}\multicolumn{1}{l|}{\diagbox{pseudo-labels}{Test Dataset}} & \multicolumn{2}{c|}{BDD-A~\cite{xia2018predicting}} & \multicolumn{2}{c|}{DR(eye)VE~\cite{alletto2016dr}} & \multicolumn{2}{c}{DADA-2000~\cite{fang2021dada}} \\ \hline
BDD-A             & \textbf{1.099±0.016} & \textbf{0.635±0.007} & 1.924±0.004 & 0.508±0.003 & \textbf{1.677±0.007} & \textbf{0.488±0.002} \\
DR(eye)VE        & 1.188±0.011 & 0.608±0.002 & 1.908±0.008 & \textbf{0.517±0.005} & 1.801±0.017 & 0.458±0.004 \\
DADA-2000        & 1.242±0.021 & 0.578±0.009 & \textbf{1.889±0.012} & 0.513±0.010 & 1.711±0.015 & 0.483±0.007 \\ \hline
\textbf{Metrics} & KLD↓         & CC↑          & KLD↓         & CC↑          & KLD↓         & CC↑          \\ \hline
\end{tabular}
\vspace{1mm}
\caption{Performance comparison of our proposed unsupervised network trained with pseudo-labels generated from various self-driving datasets (BDD-A, DR(eye)VE, DADA-2000) and then test on each benchmark. The best result is highlighted in bold.}
\label{table:three}
%%\vspace{-2mm}
\end{table*}

In the experiments, we first compare our proposed unsupervised method with other full-supervised networks on several widely-adopted datasets, \emph{i.e.,} BDD-A, DR(eye)VE, DADA-2000. Subsequently, extensive ablation studies are conducted to verify the effectiveness of each proposed component in our proposed network.

\subsection{Experimental Settings}

\noindent\textbf{Datasets.} We evaluate the performance of our proposed model on three self-driving benchmarks: BDD-A, DR(eye)VE, and DADA-2000. \textbf{BDD-A}~\cite{xia2018predicting} is an in-lab driving attention dataset consisting of 1,232 short time slices (each within 10 seconds). It contains a large amount of data from driving on various urban and rural roads. We follow its split and obtain 28k frames for training, 6k frames for validating, and 9k frames for testing. \textbf{DR(eye)VE}~\cite{alletto2016dr} is an in-car dataset that tries to maintain consistent driving conditions under controlling variables, and it contains 74 long videos in total~(each is up to 5 minutes long). We follow~\cite{alletto2016dr} and choose the last 37 videos as the test set. \textbf{DADA-2000}~\cite{fang2021dada} is another in-lab dataset and the only one including vehicle crash cases, which offers us the possibility to predict driving attention under extreme critical scenarios. This dataset contains $2000$ video clips and has over $658,746$ frames. We follow~\cite{fang2021dada} to split all videos at the ratio of 3:1:1 for training, validating, and testing. 
% Our proposed model is trained using pseudo-labels generated from the BDD-A dataset.

\noindent\textbf{Metrics.} To comprehensively evaluate our model, we utilize two common metrics, \emph{i.e.,} Kullback-Leibler divergence (KLD)~\cite{kullback1968information} as well as Pearson Correlation Coefficient (CC)~\cite{pearson1920notes}. KLD evaluates the similarity between the predicted driving attention map and the real distribution, and it is an asymmetric dissimilarity measure that penalizes false negative (FN) values more than false positive (FP) values. While CC evaluates how much the predicted driving attention map is linearly correlated with the real distribution, it is a symmetric similarity measure that penalizes equally for both FN and FP. Notice that we do not adopt the discrete metrics, such as Area Under ROC Curve ($\textit{AUC}$) and its variants ($\textit{AUC-J}, \textit{AUC-S}$), Normalized Scanpath Saliency ($\textit{NSS}$), and Information Gain ($\textit{IG}$)~\cite{bylinskii2018different} because the continuous distribution metrics is observed to be more appropriate to predict risky pixels and areas in driving scenarios~\cite{pal2020looking}.

\noindent\textbf{Compared Methods.} We compare our proposed unsupervised approach with recent fully-supervised state-of-the-art methods, including Multi-Branch~\cite{palazzi2018predicting}, HWS~\cite{xia2018predicting}, SAM~\cite{cornia2018predicting}, TASED-Net~\cite{min2019tased}, MEDIRL~\cite{baee2021medirl}, ML-Net~\cite{cornia2016deep}, UNISAL~\cite{droste2020unified}, PiCANet~\cite{liu2018picanet} and DADA~\cite{fang2021dada}.

\subsection{Implement Details}

Our proposed network is implemented using PyTorch~\cite{paszke2019pytorch}. For each dataset, we first sample both the original video frame and the gaze annotated maps to 3$HZ$, making them aligned with each other. During training, the generated pseudo-labels and the original images are resized to $224 \times 224$, and the values are normalized in the spatial dimension. Regarding the knowledge embedding strategy, we use Mask R-CNN pre-trained on the MS-COCO~\cite{lin2014microsoft} to segment important instances and fuse them with pseudo-labels. Furthermore, we set the initial learning rate of our proposed network to $0.001$, using a learning scheduler that first warm-up and then descends in a cosine fashion. Additionally, we use the Adam optimizer~\cite{loshchilov2018decoupled} ($\beta_{1}=0.9,\beta_{2}=0.999$) with the weight decay $0.0001$. Overall, we run $10$ epochs with a batch size of $32$ for training, and the training time of our proposed network is approximately 50 minutes on a single RTX 3090 GPU. While it takes about $12$~ms to infer attention regions per frame. The code will be made publicly available.

\begin{table*}
\centering
\setlength\tabcolsep{4.6mm}
\begin{tabular}{l|ll|ll|ll}
\hline
\rowcolor{gray!20}
\multicolumn{1}{c|}{} &
  \multicolumn{2}{c|}{\textbf{BDD-A}~\cite{xia2018predicting}} &
  \multicolumn{2}{c|}{\textbf{DR(eye)VE}~\cite{alletto2016dr}} &
  \multicolumn{2}{c}{\textbf{DADA-2000~\cite{fang2021dada}}} \\ \cline{2-7}
\rowcolor{gray!20}
\multicolumn{1}{c|}{\multirow{-2}{*}{\textbf{Ablated Variants}}} &
  KLD↓ &
  \multicolumn{1}{c|}{CC↑} &
  KLD↓ &
  \multicolumn{1}{c|}{CC↑} &
  KLD↓ &
  CC↑ \\ \hline
APB(unsupervised)                & 1.233 & 0.608 & 2.013 & 0.501 & 1.805 & 0.460 \\
APB+UMB                          & 1.141 & 0.622 & 1.941 & 0.510 & 1.702 & 0.480 \\
APB+UMB+non-local block          & 1.134 & 0.626 & 1.917 & 0.514 & 1.695 & 0.485 \\
\textbf{Ours}:APB+UMB+non-local block+KEB & \textbf{1.099} & \textbf{0.635} & \textbf{1.901} & \textbf{0.518} & \textbf{1.677} & \textbf{0.488} \\ \hline
\end{tabular}
\vspace{1mm}
\caption{Comparison between our proposed unsupervised model and its ablated variants. All models are trained with pseudo-labels generated from BDD-A and tested on other self-driving attention datasets (BDD-A, DR(eye)VE, DADA-2000). We ablate parts of the proposed model in each iteration until the basic APB is left alone. The basic APB is trained with unsupervised learning using pseudo-labels generated from the BDD-A training set by ML-Net. The best result is highlighted in bold.}
\label{table:ablated}
%%\vspace{-2mm}
\end{table*}

\begin{table}[!t]
\centering
\setlength\tabcolsep{7.5mm}
\begin{tabular}{l|cc}
\hline
\rowcolor{gray!20}\textbf{Pseudo-labels}       & \multicolumn{1}{c}{\textbf{KLD↓}} & \multicolumn{1}{c}{\textbf{CC↑}} \\ \hline
M        & 1.233                    & 0.608                   \\
U          & 1.246                    & 0.597                   \\
M+U & \textbf{1.099}           & \textbf{0.635}          \\
M+U+T & 1.189                    & 0.619                   \\
M+U+S           & 1.162                    & 0.621                   \\
M+U+T+S           & 1.167                    & 0.620                    \\ 
\hline
\end{tabular}
\vspace{2mm}
\caption{Comparison of different sources of pseudo-labels in the UMB on the model performance. In this table, we use the following abbreviations: M for ML-Net~\cite{cornia2016deep}, U for UNISAL~\cite{droste2020unified}, T for TASED-Net~\cite{min2019tased}, and S for SAM~\cite{cornia2018predicting}. }
\label{table:number}
\end{table}

%%The first and second rows do not use UMB but are the result of unsupervised training of APB directly using the pseudo-labels corresponding to the BDD-A datasets generated by ML-Net and UNISAL, respectively.

\begin{table}
\centering
\setlength\tabcolsep{3.9mm}
\begin{tabular}{l|cc}
\hline
\rowcolor{gray!20}\textbf{Input}                            & \textbf{KLD↓}  & \textbf{CC↑}   \\ \hline
concat (obj. \& text) & 1.126 & 0.626 \\ 
concat (obj.)          & 1.123 & 0.628 \\
single (obj. \& text, $\alpha=0.3$)            & 1.123 & 0.631 \\
single (obj., $\alpha=0.3$)            & \textbf{1.099} & \textbf{0.635} \\
\hline
\end{tabular}
\vspace{2mm}
\caption{Comparison of different strategies and types of knowledge embedding, where ``obj." refers to the masks of segmented critical objects with Mask-RCNN, ``text" refers to the masks of detected text~(\eg road signs, stop signs, etc.) with EAST in the traffic scene, and $\alpha$ means the hyper-parameter in Eq.~\ref{eqo:prior}.}
\label{table:way}
%%\vspace{-2mm}
\end{table}

\subsection{Quantitative Comparisons}

The quantitative performance of our proposed unsupervised network compared with other fully-supervised state-of-the-art models can be found in Table~\ref{table:overview}. Note that in our experiments, our unsupervised model does not utilize any ground-truth labels from self-driving datasets, but is only trained with the generated pseudo-labels with the input BDD-A training set, and then tested on each benchmark's test set. From Table~\ref{table:overview}, we can clearly observe that the proposed uncertainty network achieves competitive results compared to all fully-supervised methods and even outperforms previous fully-supervised methods in terms of the KLD metric on BDD-A and DADA-2000, and achieves the second-best w.r.t CC on BDD-A and DR(eye)VE, demonstrating the effectiveness and potential of our proposed unsupervised method.

In order to examine the transferability of these three self-driving benchmarks~(\emph{i.e.,} BDD-A, DR(eye)VE, DADA-2000), we report the results of our method trained with pseudo-labels generated in each dataset and tested on another dataset in Table~\ref{table:three}. We can find that the model trained with pseudo-labels generated from BDD-A's raw images performs the best on the test sets of two other datasets (BDD-A, DADA-2000). On the test set of the DR(eye)VE dataset, the network trained with pseudo-labels generated from DR(eye)VE's raw images performs the best on the CC metric, while the network trained with pseudo-labels generated from DADA-2000's raw images performs the best on the KLD metric indicating a superior transferability of our method. Furthermore, we discover that the images from BDD-A capture more diverse and generalized self-driving scenes, resulting in more useful and reliable pseudo-labels for our unsupervised method. Hence, our final model in this work uses the pseudo-labels generated from BDD-A. 

\subsection{Ablation Studies}
\label{subsec:abl}

\noindent\textbf{Impact of different modules.} In Table~\ref{table:ablated}, we examine each module of our proposed unsupervised model to verify their effectiveness. It can be seen that unsupervised training of APB with the pseudo-label generated from BDD-A achieves the worst performance. When we include UMB with multiple branches, the performance of the model improves significantly, far exceeding APB. Further, by adding the non-local block, we can also observe an obvious improvement. Finally, KEB brings a solid improvement to the model, making the results of our full model compatible with the state-of-the-art fully supervised models. In a word, each module in the study contributes to the final performance, while the proposed modules in this paper (UMB and KEB) contribute the most.

\begin{figure*}[!t]
    \centering
    \includegraphics[width=0.87\linewidth]{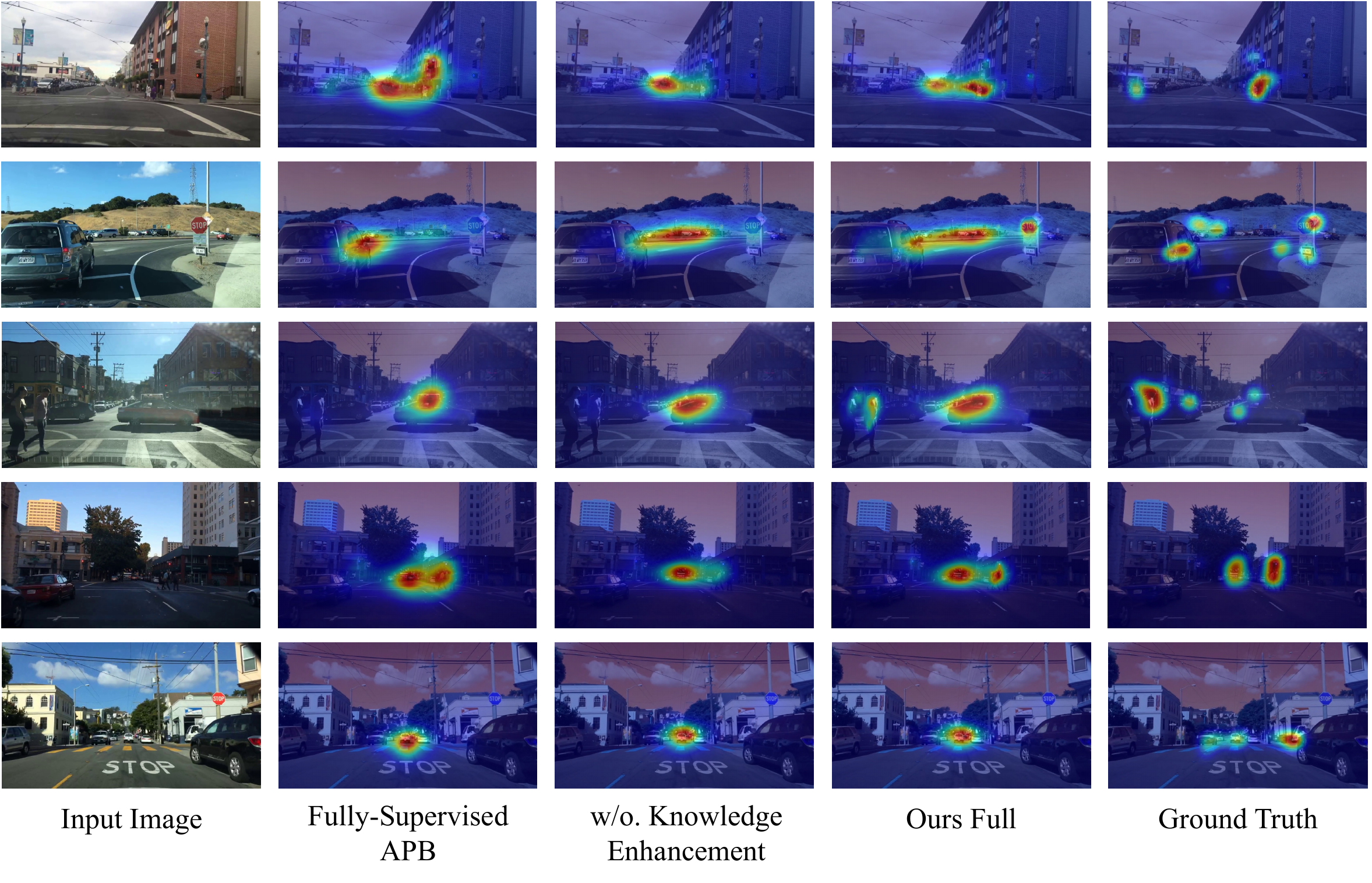}
    \caption{Visualization of the attention prediction results from different methods, \emph{i.e.,} fully-supervised APB, our method without knowledge embedding, and our full method. The results show the effectiveness of our full model in locating critical areas in the driving scene. A failure case is shown in the last row.}
    \label{fig:comparison}
    \vspace{-2mm}
\end{figure*}

\noindent{\bf Different source of pseudo-labels.} To examine the effect of different sources of pseudo-labels on the final results, we compare the performance of different pseudo-labels as is shown in Table~\ref{table:number}. The first two rows indicate the results of training with a single source pseudo-label (\eg ML-Net or UNISAL), while the third row indicates the best results of training with two sources pseudo-labels together (\emph{i.e.,} ML-Net+UNISAL) to explore uncertainty, demonstrating our UMB is able to enhance the final performance through the interaction between multiple sources of pseudo-labels. However, more than two sources of pseudo-labels result in a performance drop, as illustrated in the subsequent few lines. Therefore we choose two source pseudo-labels (ML-Net and UNISAL) in all our experiments.

\noindent{\bf Prior knowledge.} The proposed KEB in our model is used to migrate the self-driving or traffic knowledge to refine the generated pseudo-labels from the model pre-trained on the natural scenes. However, there remain problems, 1) what prior traffic knowledge should be added? 2) How to add such prior knowledge to the generated pseudo-label? Hence, we explore different ways of adding prior knowledge to add it more effectively. Here we use a pre-trained Mask R-CNN~\cite{he2017mask} to segment important traffic instances denoted as ``object'' like pedestrians and traffic lights, and we also adopt a pre-trained OCR text detection model (EAST~\cite{zhou2017east}) to segment important texts denoted as ``text'' like road signs and billboard. We can see in Table~\ref{table:way} that segmenting only important traffic instances achieve the best performance. Furthermore, we examine two different adding methods in KEB, \emph{i.e.,} combining different categories of prior knowledge with pseudo-labels by concatenation along the channel dimension denoted as ``concat'', or by operation in Eq.~\ref{eqo:prior} denoted as ``single''. As is shown in Table~\ref{table:way}, the result indicates that using the operation in Eq.~\ref{eqo:prior} works best. 

% and we hope to make them more adaptable to autonomous driving scenarios by means of adding prior knowledge to them.

\begin{table}[!t]
\centering
\setlength\tabcolsep{5.9mm}
\begin{tabular}{l|cc}
\hline
\rowcolor{gray!20}\textbf{Training strategy} & \textbf{KLD↓}  & \textbf{CC↑}   \\ \hline
fully-supervised APB  & \textbf{1.039} & \textbf{0.657} \\
semi-supervised v1   & 1.669 & 0.422 \\
semi-supervised v2   & 1.130 & 0.629 \\
unsupervised      & 1.099 & 0.635 \\ 
\hline
\end{tabular}
\vspace{1mm}
\caption{Comparing the different training paradigms, \emph{i.e.,} supervised, semi-supervised and unsupervised settings.}
\label{table:semi}
%%\vspace{-2mm}
\end{table}

\noindent{\bf Semi-supervised setting.} In addition, we also compare the semi-supervised settings following~\cite{44873} upon the same network, and the results are reported in Table~\ref{table:semi}. Specifically, we conduct two semi-supervised training schemes: 1) \textbf{Semi-supervised v1} refers to training the APB using $\frac{1}{4}$ of randomly sampled labeled data on BDD-A and then training the entire network using pseudo-labels generated from the remaining raw images; 2) \textbf{Semi-supervised v2} refers to the reversed process. However, as is shown in Table~\ref{table:semi}, we observe drastic drops in the result of the network in both Semi-supervised v1 and v2 compared with fully-supervised APB and are even inferior to our model trained in an unsupervised way. The poor performance can be explained by only using a small portion of the dataset tend to fool the model into learning a more restricted central bias, especially in self-driving. Our unsupervised method can leverage the information transferred from natural scenes by uncertainty mining, which is able to include more generalized information from non-traffic scenes to reduce bias.

%5Semi-supervised v1 refers to performing supervised training first, followed by unsupervised training. Semi-supervised v2 refers to performing unsupervised training first, followed by supervised training.

\subsection{Qualitative Results}

Figure~\ref{fig:comparison} shows visual comparisons of our model's variants on the BDD-A test set. We can observe that our full model achieves the best performance. For example, in the first row, the ground truth focuses on the pedestrians and traffic lights at the edge of the road, while the results of other methods show a strong center bias that put a lot of attention to the center of the road. Instead, our proposed model is able to reduce the central bias and assign higher attention values to the pedestrians and traffic lights in the scene which aligns with ground truth. In the second and third rows, our full model correctly focuses on the stop sign and the passing pedestrians, respectively. With an additional comparison between the third and fourth columns, we find that the proposed strategy successfully and effectively improves the final results and helps to focus on more important traffic areas of objects in the scene. To dive deep into the model's performance, a failure case is shown in the last row, where a truck tries to drive from right to left at the crossing. Our model (Ours Full) fails to focus on the truck, which is severely occluded with the nearby parked vehicles. An accurate object detection model can be further adopted to address this challenge in the future.

%%In the fourth row, the supervised model notices both road and pedestrian in the input image, while the unsupervised model without knowledge enhancement didn't pay close attention to pedestrians.

\section{Conclusion}

In this paper, we propose a novel unsupervised method for self-driving attention prediction. An uncertainty mining branch and a knowledge embedding block are introduced to generate reliable pseudo-labels and bridge the domain gap, respectively. Extensive experiments on three widely-used benchmarks demonstrate the effectiveness and superiority of our proposed method. In the future, we would incorporate the proposed method into the explainable autonomous driving control system.

{\small
\bibliographystyle{unsrt}
\bibliography{egbib}
}

\setcounter{section}{0}
\renewcommand\thesection{\Alph{section}}

This supplementary document provides further details on our proposed unsupervised self-driving attention prediction network. This includes additional information on the Knowledge Embedding Strategy in Section \ref{sec:kes}, a description of the whole algorithm in Section \ref{sec:algorithm}, a comparison of the structures of other methods in Section \ref{sec:structure}, an investigation into potential domain gaps between natural and self-driving scenes in Section \ref{sec:domain}, an explanation of the Non-local attention mechanism we used in Section \ref{sec:non}, and additional visualization examples that illustrate comparisons with other fully-supervised methods in Section \ref{sec:vis}.

%%%%%%%%% BODY TEXT
\begin{algorithm}[h]
    \small
    \caption{Knowledge Embedding Strategy}
    \label{alg:kes}
    \KwIn{Original Image $I_\textbf{input}$\; \qquad \quad Pseudo-labels $P$\; \qquad \quad Hyperparameter $\alpha$.}
    \KwOut{Knowledge-embedded pseudo-labels $\hat{P}$.}
    \BlankLine
    $O~\leftarrow~\textbf{mask-rcnn}(I_\textbf{input})$\tcc*{\scriptsize segmented instance}
    
    \ForEach{$O_\textbf{i}$ in $O$}{
        \If{is important instance mentioned in Sec~\ref{sec:kes}}{
            \ForEach{$P_\textbf{j}$ in $P$}{
                $\hat{P}_\textbf{j}~\leftarrow~P_\textbf{j}~\cdot~(O_\textbf{i}~+~\alpha)$
                \tcc*{knowledge-embedded pseudo-labels}
            }
        }
    }
\end{algorithm}

\section{Details of Knowledge Embedding Strategy}
\label{sec:kes}

\begin{figure}[t]
    \centering
    \includegraphics[width=1\linewidth]{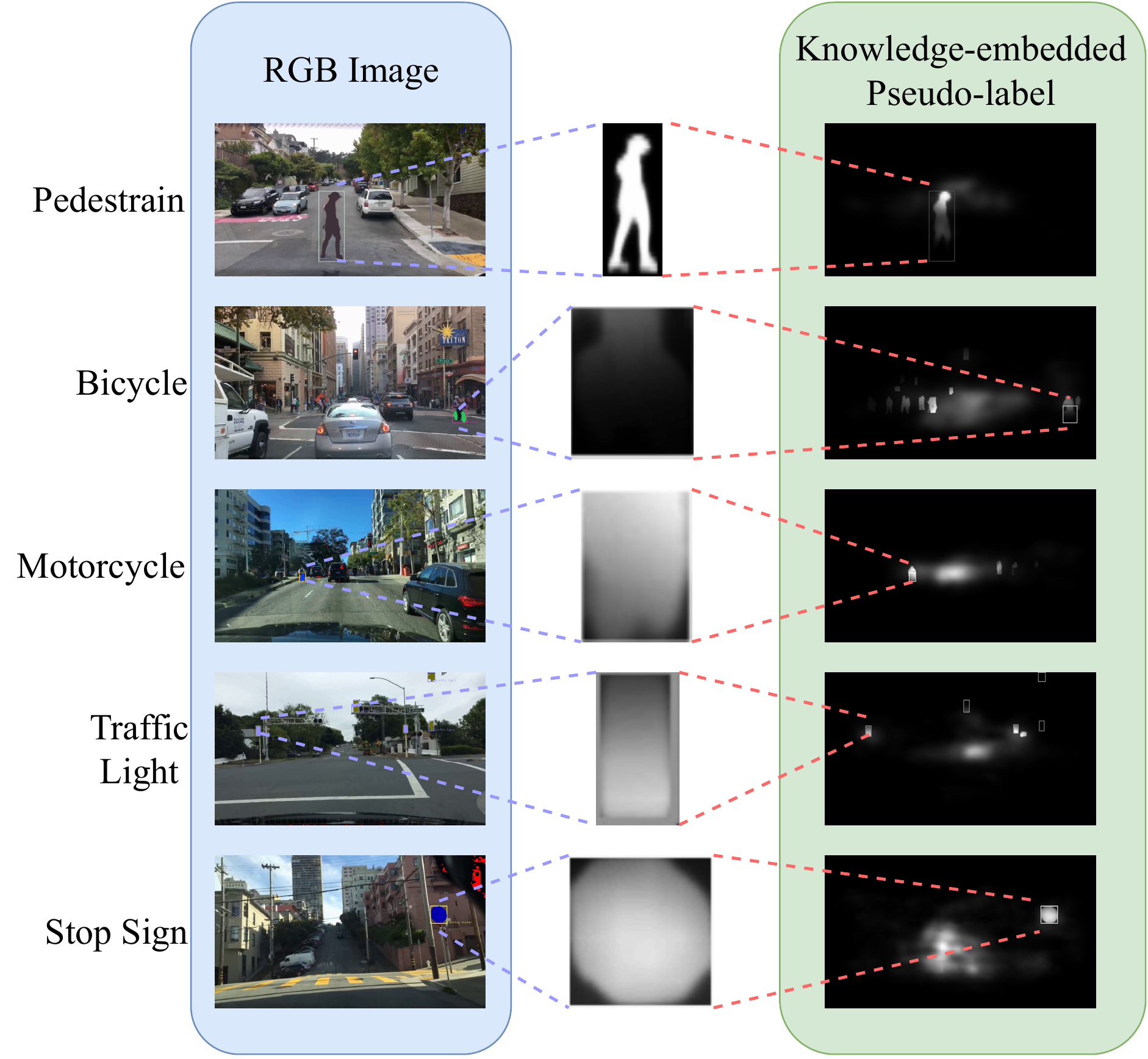}
    \caption{Illustration of our proposed knowledge embedding strategy, showing the selected important semantic clues of driving scenarios in which we perform knowledge embedding and their effects.}
    \label{fig:kes2}
\end{figure}

As mentioned in our paper, we select several representative objects in the self-driving scenario as the specific knowledge, and then embed such knowledge to refine the generated pseudo labels. Specifically, we use Mask R-CNN pre-trained on the MS-COCO~\cite{lin2014microsoft} dataset to generate the instance-level masks of the selected objects, and then merge such category-level masks into the attention map for further usage. In our method, we select \emph{pedestrian}, \emph{bicycle}, \emph{motorcycle}, \emph{traffic light}, and \emph{stop sign} among all `thing' classes as important semantics clues. As shown in Figure~\ref{fig:kes2}, we visualize the chosen important semantics of our selected driving scenarios and how they are embedded. Details of the algorithm are described in Algorithm~\ref{alg:kes}.
% We use the way mentioned in Algorithm~\ref{alg:kes} for the knowledge embedding operation.

\begin{table*}
\setlength\tabcolsep{3mm}
\centering
\begin{tabular}{|l|c|c|c|}
\hline
\multicolumn{1}{|c|}{Algorithms} & Encoder          & Decoder                             & Others         \\ \hline
ML-Net~\cite{cornia2016deep}                           & VGG              & Convolution-based                    & Learned Priors \\
SAM-VGG~\cite{cornia2018predicting} & Dilated Convolutional Network & Attentive ConvLSTM + Conv Layers            & Learned Priors \\
TASED-Net~\cite{min2019tased}                        & S3D-based blocks & Spatial-temporal decoder using C3D & -              \\
UNISAL~\cite{droste2020unified}  & MobileNet V2 encoder          & Fuse, smoothing layers and skip connections & Learned Priors \\ \hline
\end{tabular}
\caption{Architecture of four fully-supervised compared methods.}
\label{table:architec}
\end{table*}

\begin{table*}
\setlength\tabcolsep{3.6mm}
\centering
\begin{tabular}{c|ll|ll|ll|ll}
\hline
\multirow{2}{*}{\diagbox{train dataset}{test dataset}} & \multicolumn{2}{c|}{BDD-A~\cite{xia2018predicting}}      & \multicolumn{2}{c|}{DR(eye)VE~\cite{palazzi2018predicting}}  & \multicolumn{2}{c|}{DADA-2000~\cite{fang2021dada}}  & \multicolumn{2}{c}{SALICON~\cite{jiang2015salicon}}     \\ \cline{2-9} 
 &
  \multicolumn{1}{c}{KLD↓} &
  \multicolumn{1}{c|}{CC↑} &
  \multicolumn{1}{c}{KLD↓} &
  \multicolumn{1}{c|}{CC↑} &
  \multicolumn{1}{c}{KLD↓} &
  \multicolumn{1}{c|}{CC↑} &
  \multicolumn{1}{c}{KLD↓} &
  \multicolumn{1}{c}{CC↑} \\ \hline
BDD-A             & \textbf{1.036} & \textbf{0.657} & \textbf{1.870} & \textbf{0.535} & 1.824          & 0.447          & 1.584          & 0.318          \\
DADA-2000         & 1.357          & 0.543          & 2.044          & 0.484          & \textbf{1.604} & \textbf{0.504} & 1.661          & 0.351          \\
SALICON           & 2.109          & 0.287          & 2.735          & 0.277          & 2.589          & 0.247          & \textbf{0.722} & \textbf{0.552} \\ \hline
\end{tabular}
\caption{Results comparison of APB trained on different datasets and tested on another dataset. Note that BDD-A, DR(eye)VE, and DADA-2000 are self-driving benchmarks, while SALION is a natural scene dataset. The best results are highlighted in bold.}
\label{table:gap}
\end{table*}

\begin{algorithm}[t]
    \small
    \caption{State Representation of Our Method}
    \label{alg:a1}
    \KwIn{Input RGB image $I_\textbf{input}$\; \qquad \quad Pseudo-labels $P$.}
    \KwOut{Predicted attention map: $S$\; \qquad\qquad Uncertainty map: $M$.}
    \BlankLine
    \While{training}{
        $\hat{P}~\leftarrow~\textbf{KnowledgeEmbedding}(I_\textbf{input},~P)$\tcc*{embedded pseudo-labels}
        $F,~S~\leftarrow~\textbf{AttentionPrediction}(I_\textbf{input})$\tcc*{features, attention map}
        \ForEach{$F_i$ in $F$}{
            resize $F_i$ to $\frac{1}{4}$ of input image's size.
        }
        $M^0~\leftarrow~\textbf{UncertaintyBlock}(F^0, \hat{P})$\;
        $M^1~\leftarrow~\textbf{UncertaintyBlock}(F^1, M^0)$\;
        $M^2~\leftarrow~\textbf{UncertaintyBlock}(F^2, M^1)$\;
        $M~\leftarrow~\textbf{Decoder}(M^2)$\tcc*{\footnotesize uncertainty map}
        $e~\leftarrow~-\textbf{log}(M)^2$\;
        $L ~\leftarrow~\sum_n \{ L_{\mathrm{KLD}}(S, \hat{P}_n) \cdot \exp(-e_n) + \frac{1}{2}e_n \}$
        \tcc*{final loss}
    }
    \While{testing}{
        $S~\leftarrow~\textbf{AttentionPrediction}(I_\textbf{input})$.
    }
\end{algorithm}

\section{Algorithm Description}
\label{sec:algorithm}

In this part, we describe the detailed algorithm of our proposed framework in Algorithm~\ref{alg:a1}.
% Here, we exhibit the procedure of our proposed method and give some structure of methods we compared with in Section~\ref{sec:vis}. 
%The data flow of our framework is demonstrated in Algorithm~\ref{alg:a1}.
%For the network structures, we list the choices of their encoders and decoders in Table~\ref{table:architec}. Besides, we also treat the `learned priors' as an important component for comparisons.
% We show the model structures of the compared fully-supervised methods in Table~\ref{table:architec}, giving the encoder, decoder, and other important elements used (\eg, learned priors) for each model. 

\section{Compared Model Structure}
\label{sec:structure}

As listed in Table~\ref{table:architec}, we report the architecture of the chosen compared methods in our experiments, including the encoders, decoders, and `learned priors' as an important component for comparisons. Here `learned prior'~\cite{cornia2016deep,cornia2018predicting,droste2020unified} is a matrix whose size is $\frac{1}{10}$ that of the original image. All elements of `learned prior' are initialized to 1 and can be optimized during training. By dividing the final saliency map into a grid of non-overlapping cells that correspond to $\frac{1}{10}$ of the original image, `learned prior' assign weights to the attention predictions within each cell.

\begin{figure}
    \centering
    \includegraphics[width=1\linewidth]{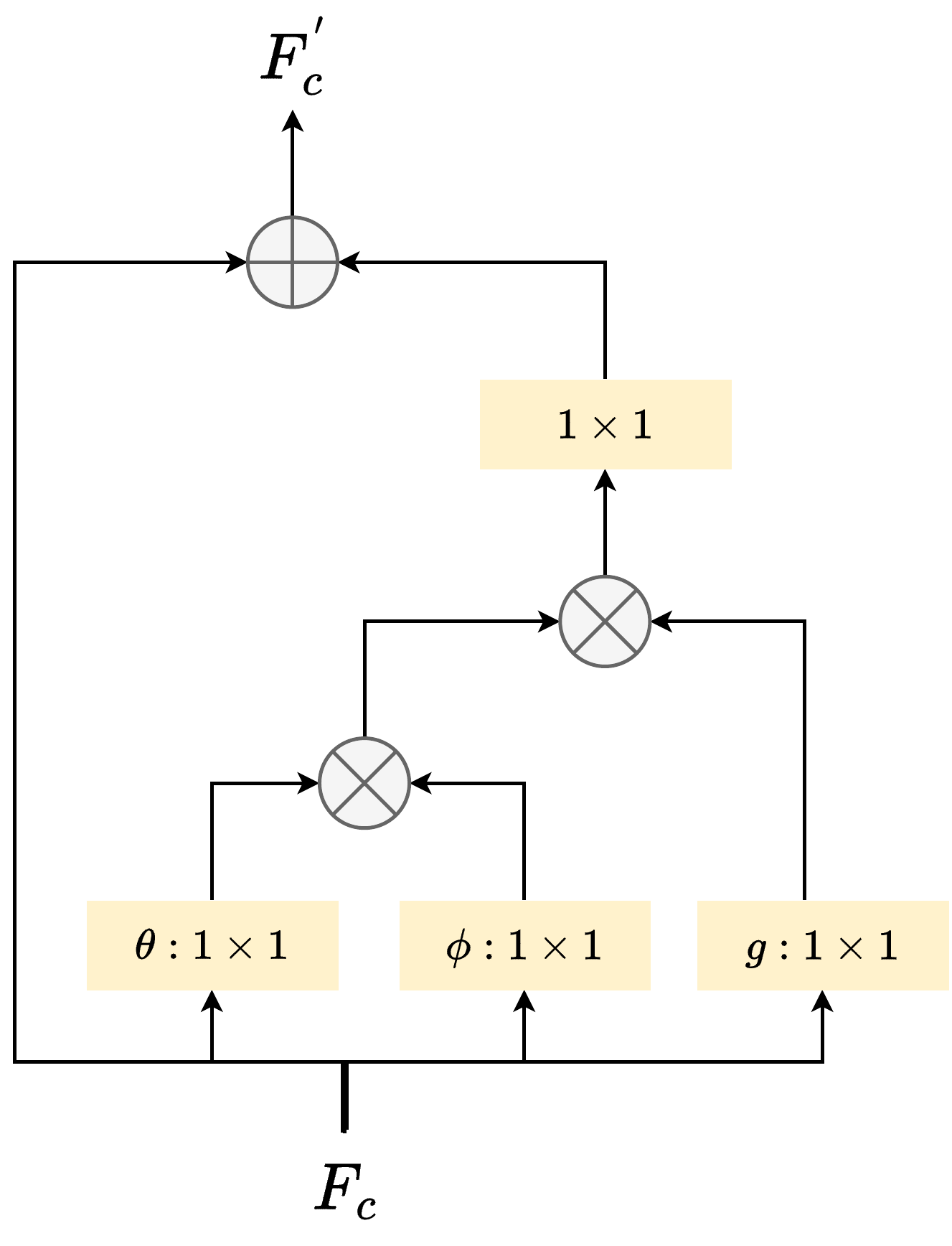}
    \caption{Illustration of the Non-local attention mechanism. The input $F_c$ yields $F_c^{'}$ of the same shape size after the self-attention calculation and the residual concatenation.}
    \label{fig:non}
\end{figure}

\begin{figure*}
    \centering
    \includegraphics[width=1\linewidth]{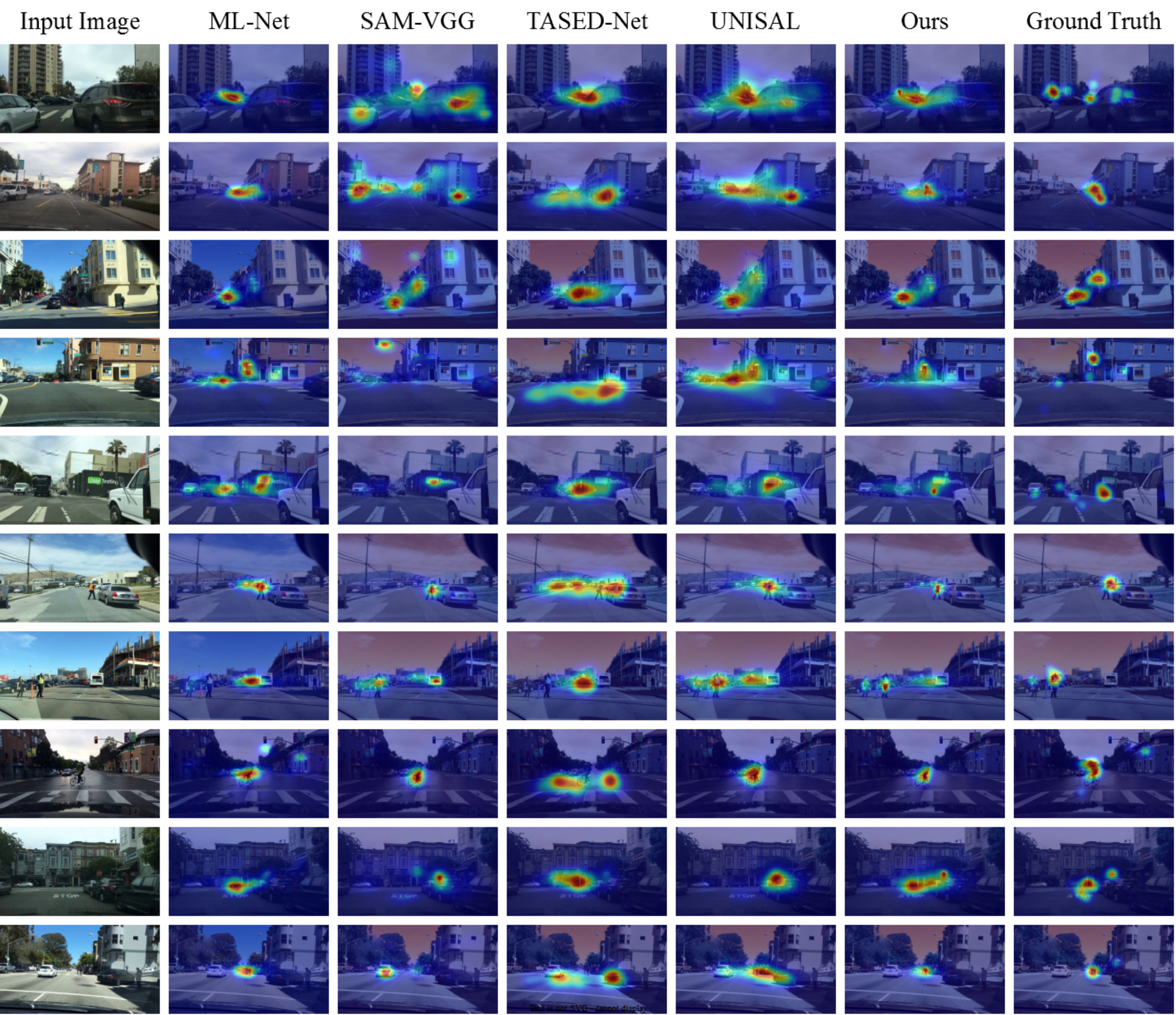}
    \caption{Comparison of our unsupervised self-driving attention prediction method with four fully-supervised methods, Part I.}
    \label{fig:part1}
\end{figure*}

\begin{figure*}
    \centering
    \includegraphics[width=1\linewidth]{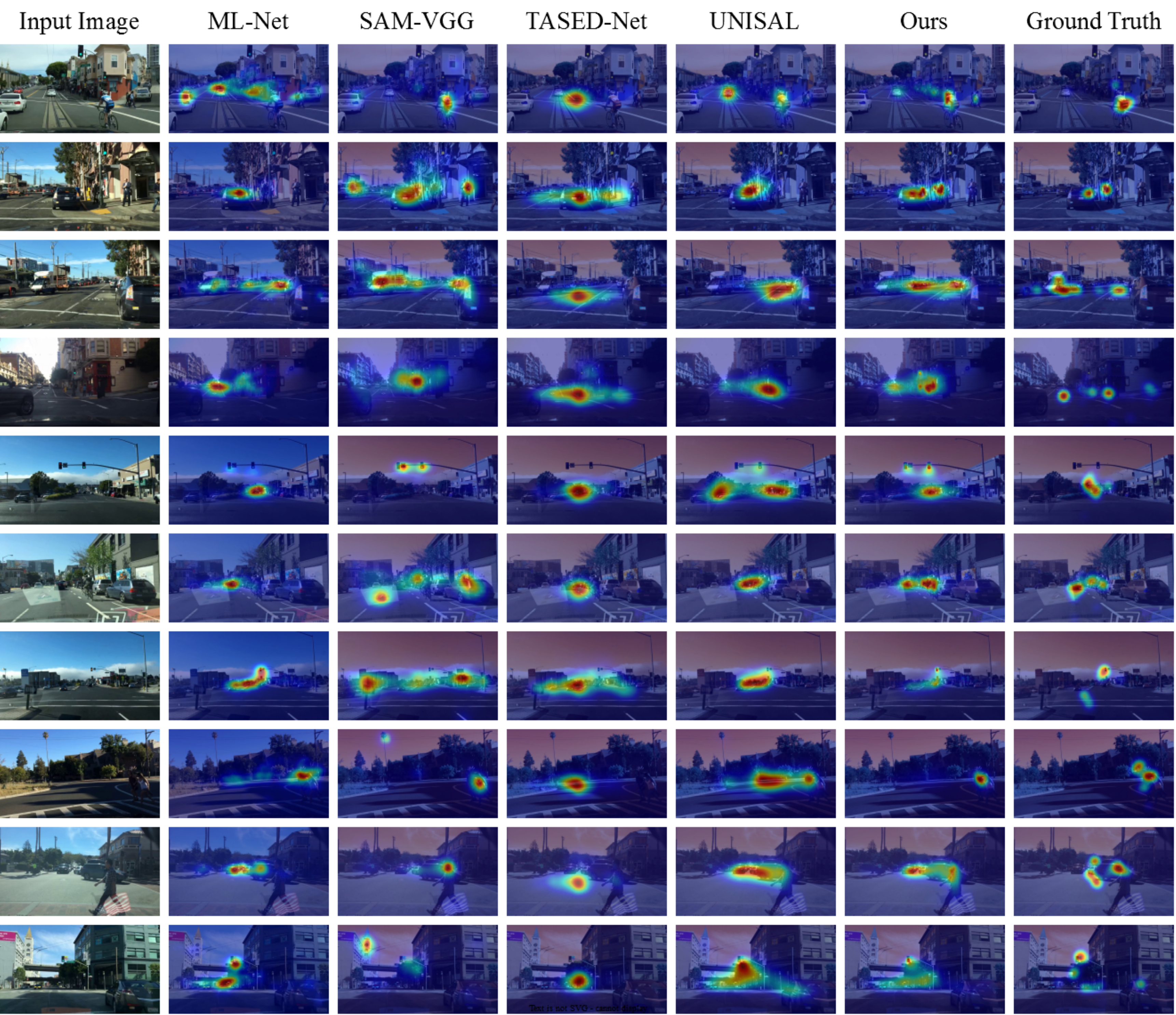}
    \caption{Comparison of our unsupervised self-driving attention prediction method with four fully-supervised methods, Part II.}
    \label{fig:part2}
\end{figure*}

\section{Domain Gap}
\label{sec:domain}

As stated in our main text, directly transferring a trained model from one domain to the other domain will bring catastrophic results due to the huge domain gap. In this section, we will conduct experiments to demonstrate it explicitly. The results are reported in Table \ref{table:gap}, indicating that there is a significant drop in performance when using the APB model trained on the first two datasets and tested on SALICON. It suggests that there is a domain gap between self-driving attention datasets and the natural scene attention dataset. Similarly, training the APB model on SALICON leads to poor performance on self-driving attention datasets. It is crucial to note that a model trained on one domain cannot be directly applied to another domain without accounting for domain differences, as this may result in poor performance.

\section{Non-local}
\label{sec:non}

As mentioned in Section 3.2 of the original text, we apply the Non-local~\cite{wang2018non} spatial attention operation in the uncertainty block to concatenate the feature ($F_c$), features passed by the Attention Predicting Block (APB), and the features of pseudo-labels, thereby finally producing the new feature ($F_c^{'}$). In this section, we present more details about the Non-local attention mechanism.

As depicted in Figure~\ref{fig:non}, the Non-local attention mechanism is formulated as a combination of self-attention and residual connection, where $\theta$, $\phi$, and $\mathit{g}$ may be analogously interpreted as $Q$, $K$, and $V$ in self-attention mechanisms, respectively. They are implemented via convolutional operations utilizing $1\times1$ convolutional kernels. The symbol $\otimes$ denotes matrix multiplication. Subsequently, the Non-local attention can be formulated as the following:
\begin{equation}
    F_{c, i}^{'} = \frac{1}{\mathit{C}(F_{c, i})} \sum_j \mathit{f}(F_{c, i}, F_{c, j}) \mathit{g}(F_{c, j})
\end{equation}
where $\mathit{C}(\cdot)$ refers to a normalization factor, in the case of the spatial attention we employ, this factor is set to the number of pixel points in a channel's features. The function $\mathit{f}(\cdot,\cdot)$ calculates the similarity between any two points, while $\mathit{g}(\cdot)$ computes the eigenvector of a single point. The subscripts i and j denote a particular pixel location within a given channel's feature map.

\section{Additional Visualization Examples}
\label{sec:vis}

We provide more visualized examples of the comparison between our proposed unsupervised self-driving attention prediction method and several fully-supervised state-of-the-art methods in Fig~\ref{fig:part1} and Fig~\ref{fig:part2}, which consistently show the effectiveness and robustness of our proposed method. Note that input scenes are selected randomly from the validation sets in the BDD-A benchmark.
% We choose some critical scenes to show the effectiveness of our model and show our model's grasp of the important semantics of critical scenes. 

\end{document}